\documentclass{article} 
\usepackage{iclr2022_conference,times}

\usepackage{amsmath,amsfonts,bm}

\def\eqref#1{equation~\ref{#1}}

\def\1{\bm{1}}

\DeclareMathAlphabet{\mathsfit}{\encodingdefault}{\sfdefault}{m}{sl}
\SetMathAlphabet{\mathsfit}{bold}{\encodingdefault}{\sfdefault}{bx}{n}

\usepackage{hyperref}
\usepackage{url}

\usepackage{graphicx}
\usepackage{booktabs}
\usepackage{tabularx}
\newcolumntype{Y}{>{\centering\arraybackslash}X}

\usepackage[linesnumbered,algoruled,boxed,lined]{algorithm2e}
\makeatletter 
\g@addto@macro{\@algocf@init}{\SetKwInOut{Require}{Require}}

\makeatletter
\DeclareRobustCommand\onedot{\futurelet\@let@token\@onedot}
\def\@onedot{\ifx\@let@token.\else.\null\fi\xspace}

\makeatother

\newcommand{\MainMethod}{Wormhole MAML}

\title{Wormhole MAML: Meta-Learning in Glued Parameter Space}

\author{Chih-Jung Tracy Chang\thanks{Equal contribution. Listed alphabetically.}, Yuan Gao\footnotemark[1] \& Beicheng Lou\footnotemark[1] \\
Stanford University \\
\texttt{\{chihjung,y1gao,lbc45123\}@stanford.edu} \\
}

\iclrfinalcopy 
\begin{document}

\maketitle

\begin{abstract}


In this paper, we introduce a novel variation of model-agnostic meta-learning, where an extra multiplicative parameter is introduced in the inner-loop adaptation. 
Our variation creates a shortcut in the parameter space for the inner-loop adaptation and increases model expressivity in a highly controllable manner.
We show both theoretically and numerically that our variation alleviates the problem of conflicting gradients and improves training dynamics.
We conduct experiments on 3 distinctive problems, including
a toy classification problem for threshold comparison, a regression problem for wavelet transform, and a classification problem on MNIST.
We also discuss ways to generalize our method to a broader class of problems.
\end{abstract}

\section{Introduction}
Developing artificial intelligence that can learn to learn has been a long-sought-after goal.
On the other hand, while deep learning has achieved impressive results on various problems in computer vision, natural language processing, and reinforcement learning, the cost of training sophisticated models for new tasks from scratch has also become tremendous. As a result, multi-task and meta-learning have emerged in recent years as a promising approach that aims to discover shared structure across tasks so that individual ones can be solved more efficiently~\cite{finn2017model}. The training process produces a meta-model that can achieve good performance on individual tasks upon fine-tuning. 

Various architectures and algorithms have been developed for multi-task and meta-learning.
One major variability lies in the choice of what to share in the joint model. 
For example, in prototypical network~\cite{snell2017prototypical}, all learned parameters are shared across tasks, while the data for different tasks are encoded into a low-dimensional representation.
Similarly, in recurrent models~\cite{duan2016rl}, one model is shared across tasks and task-specific information is implicitly encoded in the state vector.
In contrast, in model-agnostic meta-learning (MAML)~\cite{finn2017model}, the parameters for each layer of the network are allowed to be varied during the adaptation.
In fast context adaptation for meta-learning~\cite{zintgraf2019fast}, one explicitly separates parameters into shared ones and task-specific ones.

One common challenge in these approaches is training efficiency. Memory-based methods using recurrent neural networks tend to omit inductive bias and easily lead to non-converging behavior~\cite{flennerhag2019meta}. Gradient-based methods are typically restricted to few-shot learning due to the need for gradient tracking during adaptation~\cite{flennerhag2019meta}. Otherwise, they may be prone to overfitting~\cite{rajendran2020meta}. In general, the extra hierarchy in the update process requires a more delicate design of the update rule~\cite{li2017meta, gradient-clipping, lou2021}. 

The problem with training efficiency is especially notable when data is relatively scarce, such as in reinforcement learning settings~\cite{yu2020gradient}. The joint training of multiple tasks tends to introduce undesired interference between task gradients and negative transfer across tasks, where the adaptation performance is worse than single-task learning. These conflicting gradients could lead to a highly inefficient trajectory of gradient updates and slow down the optimization. 

\begin{figure}[h]
\begin{center}
\includegraphics[width=0.49\linewidth]{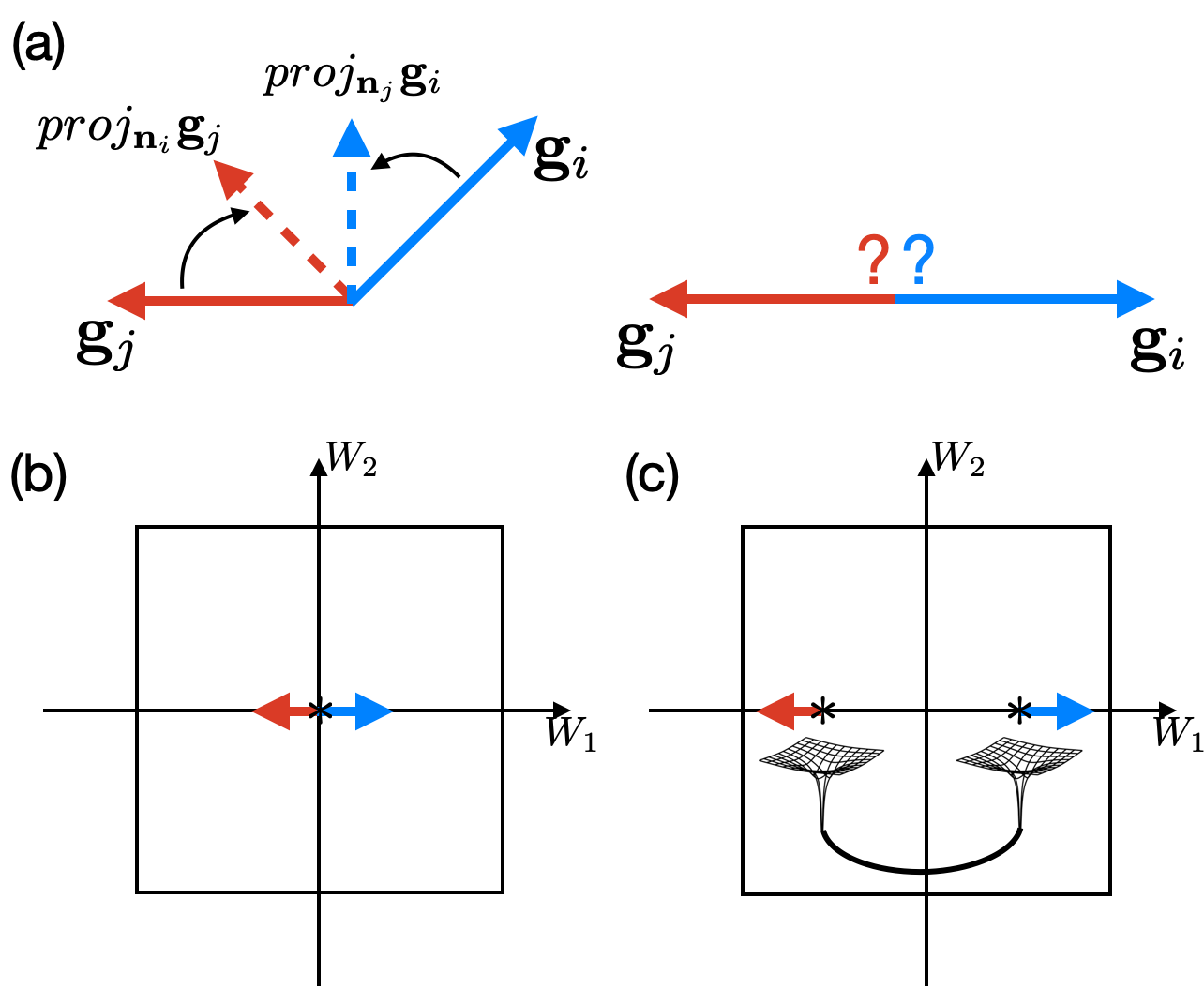}
\end{center}
\caption{\textbf{Illustration of the problem of conflicting gradients.} (a, b) Illustration of gradient surgery and its limitation. (c) Illustration of "wormholes" in the parameter space that resolves the problem of conflicting gradients.}
\label{fig:illus_gradient}
\end{figure}

One intuitive approach to tackle conflicting gradients in meta-learning is gradient surgery~\cite{yu2020gradient}, where each task gradient is projected to the normal plane of other conflicting gradients for updates. However, this approach is only applicable when the gradient conflict is not too bad, so that a significant component is projected onto the normal plane. As illustrated in Fig.~\ref{fig:illus_gradient}a, when the conflicting gradients are nearly opposite, the projection is zero and completely dominated by noise. The model is therefore trapped in the parameter space, as illustrated in Fig.~\ref{fig:illus_gradient}b. While it is still possible for the inner-loop adaptation to reach task-specific optima, there is hardly any inductive bias learned in the shared parameters in this case.

In this paper, we propose a variant architecture of MAML, with an extra multiplicative parameter applied to the network weights that is both task-specific and highly regularized. The extra multiplicative parameter creates a shortcut in the parameter space and effectively glues certain parts together, analogous to the concept of \emph{wormhole}. Such shortcuts allow the shared parameters to align their gradients across tasks, as illustrated in Fig.~\ref{fig:illus_gradient}c, free from the problem of conflicting gradients.





\section{Wormhole MAML}
In MAML, the meta-network parameter $\theta$ is learned through an outer loop and the task-specific parameter $\phi_i$ is learned during the inner loop.
For simplicity, let $\phi$, $\theta$ directly represent the weight matrices in the neural network, i.e. $W_\phi=\phi$, $W_\theta = \theta$. Then the inner-loop update of MAML can be written as:
\begin{equation}
    \phi_i=\theta+\Delta_i(\theta)
\end{equation} 
where $\Delta_i(\theta)$ is the aggregated steps in the gradient descent direction. In the case of single-step update, $ \Delta_i(\theta) = \alpha \cdot \nabla \mathcal{L}_{{\mathcal{T}}_{i}}(\theta) $.

In Wormhole MAML, the inner-loop update is changed to:
\begin{equation}\phi_i=C_i\theta+\Delta_i(\theta)\end{equation}
where $\Delta_i(\theta)$ is still the aggregated steps that ultimately depend on $\theta$ only, but now it also has a partial dependence on $C_i$.

This extra multiplicative parameter in the inner-loop update allows some flexibility to incorporate our prior knowledge for better training efficiency. For example, to tackle the problem of conflicting gradients illustrated in Fig.~\ref{fig:illus_gradient}a, if $C_i=1$ and $C_j=-1$ for the two contradicting tasks $i$ and $j$, then the gradient on $\theta$ can be highly aligned, leading to an efficient trajectory for the outer-loop parameters.

As in MAML, the algorithm could have some variations, especially in terms of update rules for the multiplicative parameter $C_i$.
In the simplest case, we can add a step between the outer-loop update and the inner-loop update, and the algorithm is:
\begin{algorithm}[H]
\Require{$p(\mathcal{T})$: distribution over tasks}
\Require{$\alpha, \beta, \gamma$: step size hyper-parameters}
Randomly initialize $\theta$\;
\While{not done}{
    Sample batch of tasks $\mathcal{T}_i \sim p(\mathcal{T})$\;
    \For{all $\mathcal{T}_i$}{
        \textcolor{red}{Initialize $C_i$\;}
        \textcolor{red}{Evaluate $\nabla \mathcal{L}_{\mathcal{T}_i}(\theta, C_i)$ with respect to $K$ examples\;}
        \textcolor{red}{Update $C_i \leftarrow C_i - \gamma \nabla_{C_i} \sum_{\mathcal{T}_i \sim p(\mathcal{T})} \mathcal{L}_{\mathcal{T}_i}(\phi_i, C_i)$\;}
        \textcolor{red}{Compute adapted parameters with gradient descent: $\phi_i \leftarrow C_i \theta + \nabla \mathcal{L}_{\mathcal{T}_i}(\theta, C_i)$\;}
    }
    Update $\theta \leftarrow \theta - \beta \nabla_\theta \sum_{\mathcal{T}_i \sim p(\mathcal{T})} \mathcal{L}_{\mathcal{T}_i}(\phi_i, C_i)$\;
    
}
\caption{\MainMethod}\label{alg:ours}
\end{algorithm}
where the part that differs from conventional MAML is marked in red. The initialization of $C_i$ could have multiple variations depending on the specific parameterization.
For example, $C_i$ can be parametrized in one of the following forms:
\begin{align}
&C_i = c, \; c \in \mathbb{R}^1 \label{eq:c1}\\
&C_i = \tanh(t), \; t \in \mathbb{R}^1 \label{eq:c2}\\
&C_i = diag(C), \; C \in \mathbb{R}^{d_\theta} \label{eq:c3}\\
&C_i = diag(C), \; C \in \mathbb{R}^{L} \label{eq:c4}\\
&C_i = C, \;  C \in \mathbb{R}^{d_\theta \times d_\theta}, Rank(C) < threshold
\end{align}
where $d^\theta$ is the dimension of the meta-network parameter and $L$ is the number of layers.
This paper mainly focuses on the case of Eq.~\ref{eq:c1}, \ref{eq:c2},\ref{eq:c4}. The initialization of $C_i$ is typically around $1$. Both the updates for $C_i$ and $\phi_i$ can be more than a single step. In particular, for $C_i=tanh(t)$ parametrization, we typically let $C_i$ converge towards $\pm 1$.

\begin{figure}[t]
\begin{center}
\includegraphics[width=0.59\linewidth]{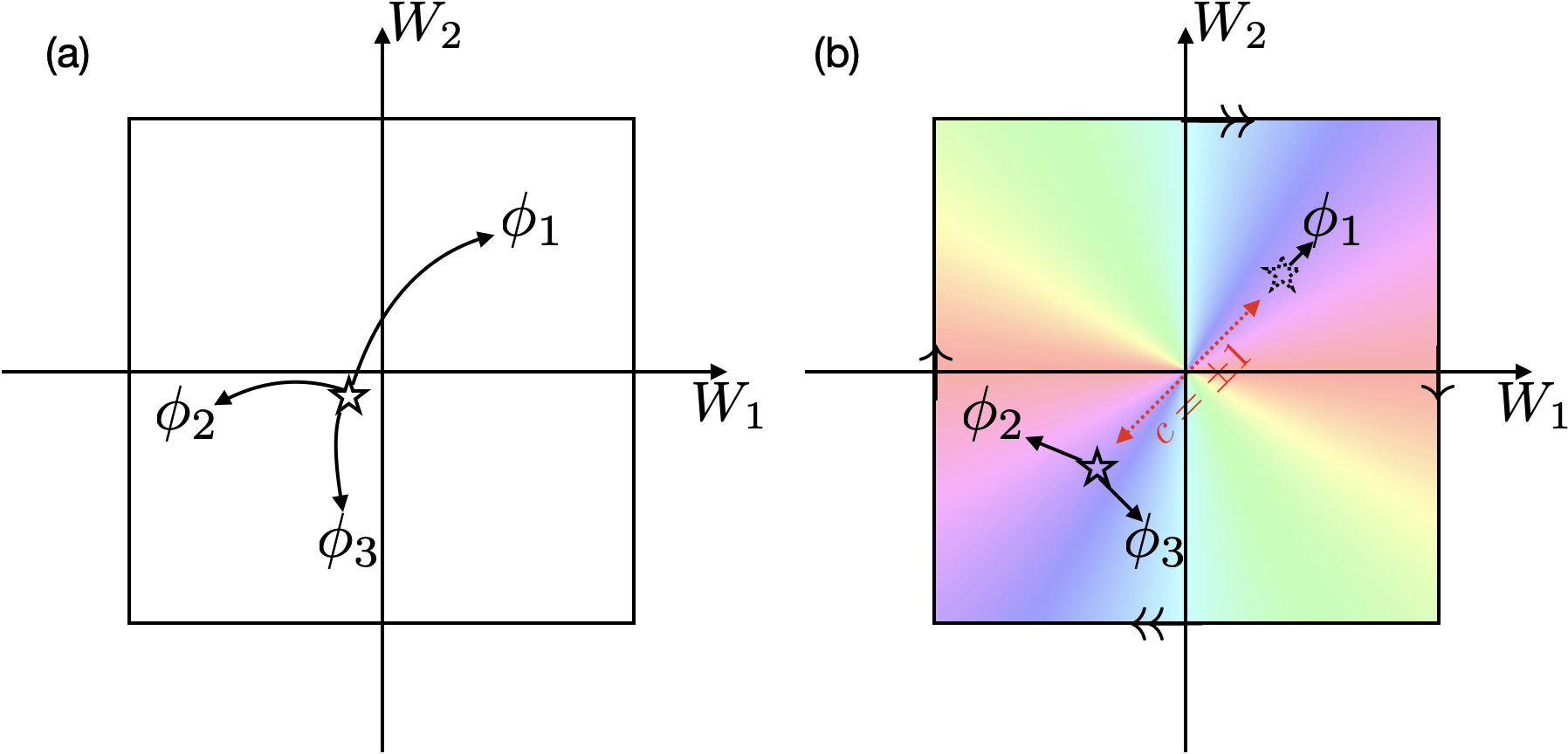}
\end{center}
\caption{(a) Illustration of vanilla MAML's optimum. (b) Illustration of wormhole MAML's optimum.}
\label{fig:illus_optimun}
\end{figure}

The mechanism of Wormhole MAML is summarized in Fig.~\ref{fig:illus_optimun}:
In conventional (vanilla) MAML, the optimum in the parameter space for the meta-network need to be near the task-specific optima, so that all task-specific optima are reachable within the steps allowed during inner-loop adaption, as illustrated in Fig.~\ref{fig:illus_optimun}a.
In many cases, either due to restrictions on computational resources or data availability, the inner-loop adaption is not allowed to have long trajectories and it is highly desirable for the meta-network optimum to be closer to the task-specific optima.

In Wormhole MAML, the choice of $C_i=c, \; c\in \mathbb{R}^1$ effectively glues the parameter space in a way as indicated by the color and the arrows on the boundary, where each color represents the parameter space reachable by tuning $C_i$ alone. When $c=\pm 1$, $c \theta$ can be initialized at either of the two-star locations depending on the task, which greatly reduces the adaptation needed in $\Delta_i(\theta)$ for $c \theta + \Delta_i(\theta)$ to reach the task-specific optima.
Here we want to encourage $C_i$ towards $\pm 1$, potentially with $\tanh(t)$ parameterization, since we don't want $C_i$ to absorb all the inner-loop tunability of $\Delta_i$, or $C_i$ near $0$ causing bad dynamics.

\section{Related Works}


The architecture in our work could interpolate various other approaches.

When the extra multiplicative parameter is shared across weights and not binarized, our algorithm effectively encourages weights to vary in coordination during the adaptation phase, thus maintaining a high correlation. This is conceptually similar to the idea of correlation alignment for unsupervised domain adaption (CORAL)~\cite{sun2016deep}, where correlation in data distribution is learned. More specifically, CORAL aims to align second-order statistics of source and target distribution. Despite the conceptual similarity, CORAL aims at a completely different problem of domain adaptation, whereas our approach is promoting a certain weight correlation in this setting.

When the extra multiplicative parameter is selectively frozen, our algorithm is similar to fast context adaptation (CAVIA)~\cite{zintgraf2019fast}, which explicitly separates parameters into shared ones and task-specific ones. When the additive part of the inner-loop adaption is highly constrained, the multiplicative part is the major source of model expressivity. In that case, our selection also separates weights into relatively more shared ones and relatively more task-specific ones. The main difference is that our approach achieves it softly and potentially leverages information during the training process, or even in the query data batch.

In the case where the multiplicative parameter is softly binarized with $\tanh$, potentially under some annealing scheme, the effect of the multiplicative parameter can be viewed as introducing extra variation in the learning rate of the network weights, which is similar to the idea of meta-SGD~\cite{li2017meta}, where the learning rate is also learnable. It is also similar to MT-net~\cite{mt-net}, where task-specific adaptations are restricted to a subspace. Despite the conceptual similarity, our approach has a drastically different update rule and offers more flexible control over the training dynamics. 

\section{Experiments}

To verify the effectiveness of our proposed approach, we compare Wormhole MAML with the vanilla MAML~\cite{finn2017model} in three different meta-learning tasks: \textit{Avg-Threshold}, \textit{Wavelet Transform}, and \textit{the MNIST dataset}~\cite{lecun1998mnist}. The quantitative comparison results of the three tasks can be found in Table~\ref{tab:evaluation}.

\begin{table}[t]
\begin{center}
\caption{Evaluation results on the Avg-Threshold, Wavelet Transform and MNIST task.}
\label{tab:evaluation}
\scalebox{0.87}{
\begin{tabularx}{1.2\textwidth}{l *{10}{Y}}
\toprule
 & \multicolumn{3}{c}{Avg-Threshold} & \multicolumn{3}{c}{Wavelet Transform} & \multicolumn{3}{c}{MNIST} \\
\cmidrule(lr){2-4} \cmidrule(lr){5-7} \cmidrule(lr){8-10}
 & 1-step & 2-step & 5-step & 1-step & 2-step & 5-step & 1-step & 2-step & 5-step \\
\midrule
 Vanilla~\cite{finn2017model} & 5.60 & 5.08 & 3.33 & 7.14 & 7.63 & 7.49 & \textbf{0.500} & 0.463 & 0.312 \\
 \textbf{Wormhole 1} & \textbf{1.56} & \textbf{1.41} & \textbf{1.69} & \textbf{5.00} & \textbf{6.05} & \textbf{5.36} & \textbf{0.500} & \textbf{0.141} & \textbf{0.009} \\
 \textbf{Wormhole d} & 2.53 & 2.42 & 2.17 & - & - & - & - & - & - \\
\bottomrule
\end{tabularx}}
\end{center}
\end{table}

\subsection{Avg-Threshold}

\paragraph{Problem Formulation}
We first consider a toy meta-learning problem of telling whether the average of an array is greater or less than a certain threshold. More specifically, the dataset has input domain $\mathcal{X} = [0, 1]^d$ and output domain $\mathcal{Y} = \{0, 1\}$. In each task, given a task-specific threshold $\tau \in [0, 1]$ and sign $s \in \{-1, 1\}$, the goal is for the model to learn a mapping $\phi$ such that $\phi(x) = 1$ if $s * avg(x) > s * \tau$ and 0 otherwise.

\paragraph{Implementation Details}
We conduct the experiments by training 150 epochs with a task batch size of 10. We set the input dimension $d$ to 5 and use a single linear layer as the meta-model $\theta$. We use SGD with a learning rate of 0.1 as our optimizer. The loss function is the cross-entropy error between the predicted and ground-truth label. Two types of regularization on $C_i$ are considered in the experiments: $C_i = \tanh(t), \; t \in \mathbb{R}^1$ and $C_i = diag(C), \; C \in \mathbb{R}^{d_\theta}$. We consider 1-, 2-, and 5-step inner loop adaptation in the experiments. The adaptation learning rate is set to 1.0.

\paragraph{Results and Analysis}
We report the cross-entropy error of each experiment in Table~\ref{tab:evaluation}. Wormhole MAML significantly improves over the vanilla MAML~\cite{finn2017model} by reducing $>2$ cross-entropy error in most of the settings. It is worth noting that the advantage of using wormhole parameters is still observable even in the setting of 5-step inner-loop adaptation. 
We also note that the $\mathit{tanh}$ regularization on $C_i$ results in lower error than the $\mathit{diag}$ regularization, indicating strong regularization is beneficial to simple meta-learning tasks. 

\begin{figure}[t]
\begin{center}
\includegraphics[width=1.0\linewidth]{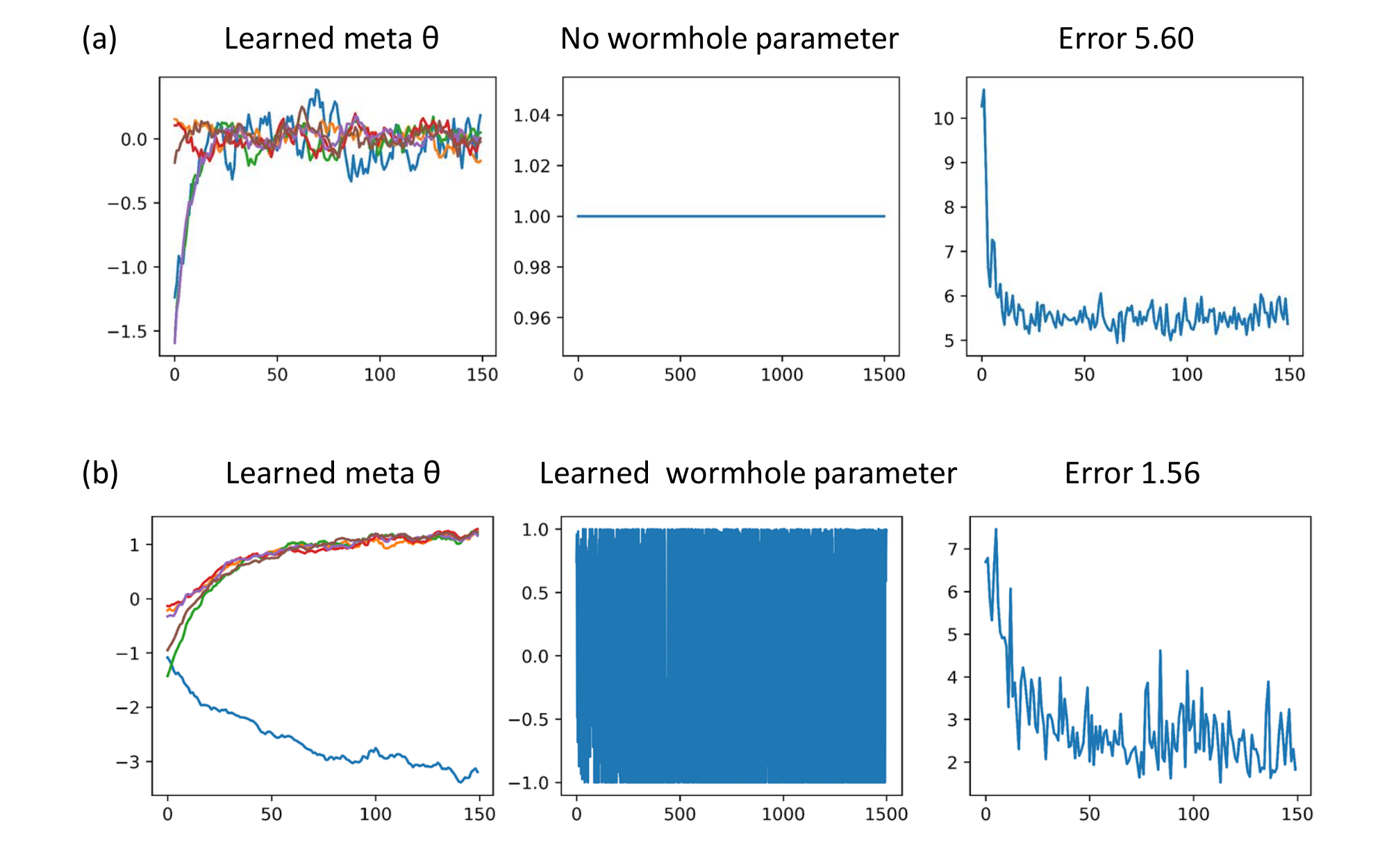}
\end{center}
\caption{\textbf{Parameter and loss curve analysis on the Avg-Threshold task.} (a) Without the adaptation of the wormhole parameter, the learned meta parameter $\theta$ shrinks toward 0. (b) The learned wormhole parameter serves as a "sign changer" in the inner loop adaptation.}
\label{fig:loss_avgthres}
\end{figure}

To further examine the training behavior under the effect of wormhole parameters, we show the parameter and loss curve analysis in Fig.~\ref{fig:loss_avgthres}. Without the adaptation of wormhole parameters (Fig.~\ref{fig:loss_avgthres}a), the learned meta-model $\theta$ shrinks toward 0 for the sake of quick 1-step adaptation to +1 or -1 sign in each task. On the other hand, by adding a wormhole parameter $C_i = \tanh(t), \; t \in \mathbb{R}^1$ (Fig.~\ref{fig:loss_avgthres}b), $\theta$ is able to learn the meaningful weight that represents averaging and thresholding, with $C_i$ acts as a "sign changer" in the inner-loop adaptation, leading to faster and lower meta-training convergence.

We note that the benefits in Wormhole MAML largely depend on the capability to find correct values for the extra multiplicative parameter $C_i$ easily and stably. In this case, the choice of $C_i$ is merely a scalar for the linear network, and is expected to be highly stable given its regularized nature.

We can characterize the stability by analytically examining the asymptotic behavior of $c$.
Without loss of generality, assume positive sign $s=1$. The loss over a batch of data is:
\begin{equation}
    L(\mathbf{x},\mathbf{y}) = -\sum_{y_i=1}\log(\hat{y}_i) -\sum_{y_i=0}\log(1-\hat{y}_i)
\end{equation}
where $\hat{y}_i = \sigma(c \mathbf{W}^T \mathbf{x}_i)$ is the predicted label. For simplicity, let $z_i = \mathbf{W}^T \mathbf{x}_i$. The partial gradient of $c$ is:
\begin{equation}
    \frac{\partial L(\mathbf{x},\mathbf{y})}{\partial c} = -\sum_{y_i=1}(1-\sigma(c z_i))z_i -\sum_{y_i=0}\sigma(c z_i) z_i
\end{equation}

To find $c^*$ such that $\frac{\partial L(\mathbf{x},\mathbf{y})}{\partial c} |_{c=c^*}=0$, consider the $|c|\ll 1$ case, where we can perform a Taylor expansion:
\begin{equation}
    \sigma(cz_i) \approx cz_i
\end{equation}
Therefore the loss function gradient w.r.t. $c$ can be approximated as:
\begin{align}
    \frac{\partial L(\mathbf{x},\mathbf{y})}{\partial c} &\approx
    -\sum_{y_i=1}(1-c z_i)z_i -\sum_{y_i=0}(c z_i) z_i \\
    &=-\sum_{i} I[z_i + \Delta \tau > 0] (1-cz_i) z_i - I[z_i + \Delta \tau < 0]c z_i^2
\end{align}
where $\Delta \tau = \tau^* - \tau$ is distance to the classification boundary and $I[\cdot]$ denotes the indicator function such that $y_i=I[\mathbf{W}^T \mathbf{x_i} + \tau^*>0]$.

Setting $\frac{\partial L(\mathbf{x},\mathbf{y})}{\partial c}$ to $0$, we get:
\begin{equation}
    c^* = \frac{\sum_i I[z_i+\Delta t>0]z_i}{\sum_i z_i^2}
    \label{eq:cstar}
\end{equation}
which ideally should be $1$.

From Eq.~\ref{eq:cstar}, we see that $t$ should be near $t^*$, on the scale of $std(\mathbf{x}_i)$. If $t\ll t^*$, we have $c\propto t$, or if $t\gg t^*$, we may get $c=0$. These two cases correspond to highly unbalanced data batch and are both undesirable, leading to unstable convergence of $c$. In the other cases especially when $|t-t^*|\ll std(\mathbf{x}_i)$, we have $c$ close to $1$ as desired.

In conclusion, the optimality of $c$ requires a balanced data batch, which is reasonable, as well as relatively accurate model parameters ($\tau$ and $W$ in this case). Therefore, it may be worth contemplating a scheme where $c$ and $W$ are co-adapted but at different rates, potentially with annealing that encourages $c$ to converge faster towards $\pm 1$.

\subsection{Wavelet Transform}
\begin{figure}[h]
\begin{center}
\includegraphics[width=0.85\linewidth]{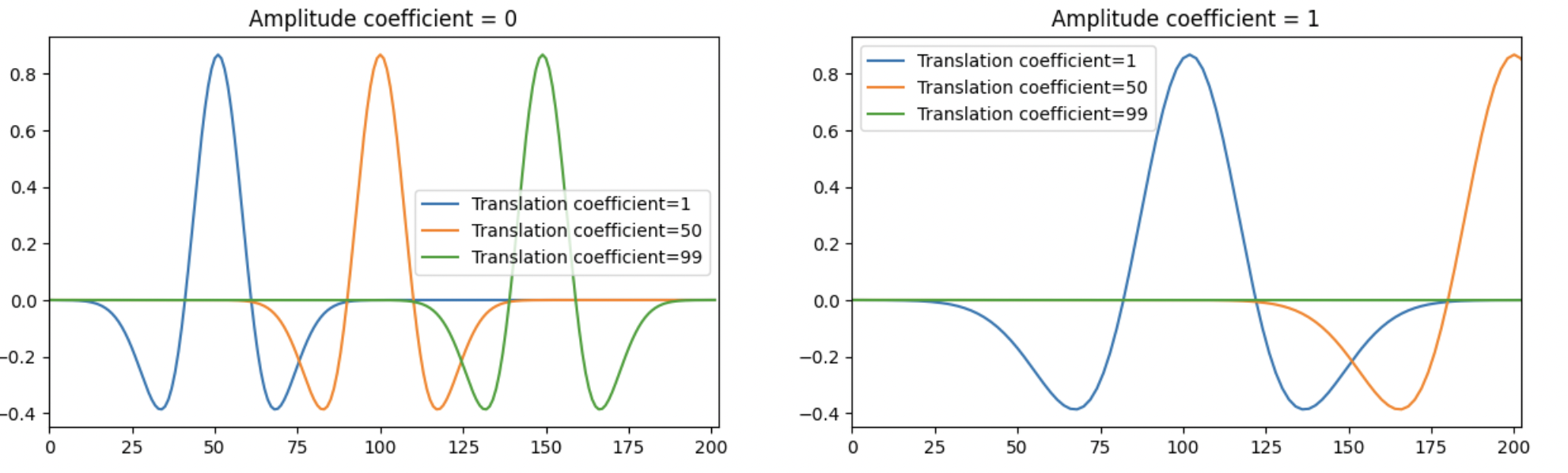}
\end{center}
\caption{Illustration of the Mexican hat wavelet transforms.}
\label{fig:illus_wavelet}
\end{figure}
\paragraph{Problem Formulation}
Another toy example to consider is to approximate wavelet functions from their dot-product with different vectors. Specifically, we consider a set of Mexican Hat wavelets $\mathcal{M}$ with different translations, as illustrated in Fig.~\ref{fig:illus_wavelet}. For each task, we sample $sign \sim \{-1,1\}, x \sim uniform (0,1)^{n}, m \sim \mathcal{M}$, where $m$ is supported on a discrete set of $n$ points. The desired filter for the wavelet transform is $f=sign*m$. The true label is $y = x \cdot f$. Our goal is to learn a $w$ and a $C$ minimizing $||x \cdot (Cw) - y||^2$. 

\paragraph{Implementation Details}
In this experiment we only consider $C_i = \tanh(t), \; t \in \mathbb{R}^1$. Unless otherwise mentioned, we adopt 1500 epochs for training, 5 inner loops for $C_i$, and 1, 2, 5 inner loops for $w$. We use 10 shots per task, and a batch size of 10 tasks. For the learning rate, the inner loop learning rate for $w$ is $5e-2$ and for $c$ is $1$, and the learning rate for the outer loop is $1e-3$.

\begin{figure}[t]
\begin{center}
\includegraphics[width=1.0\linewidth]{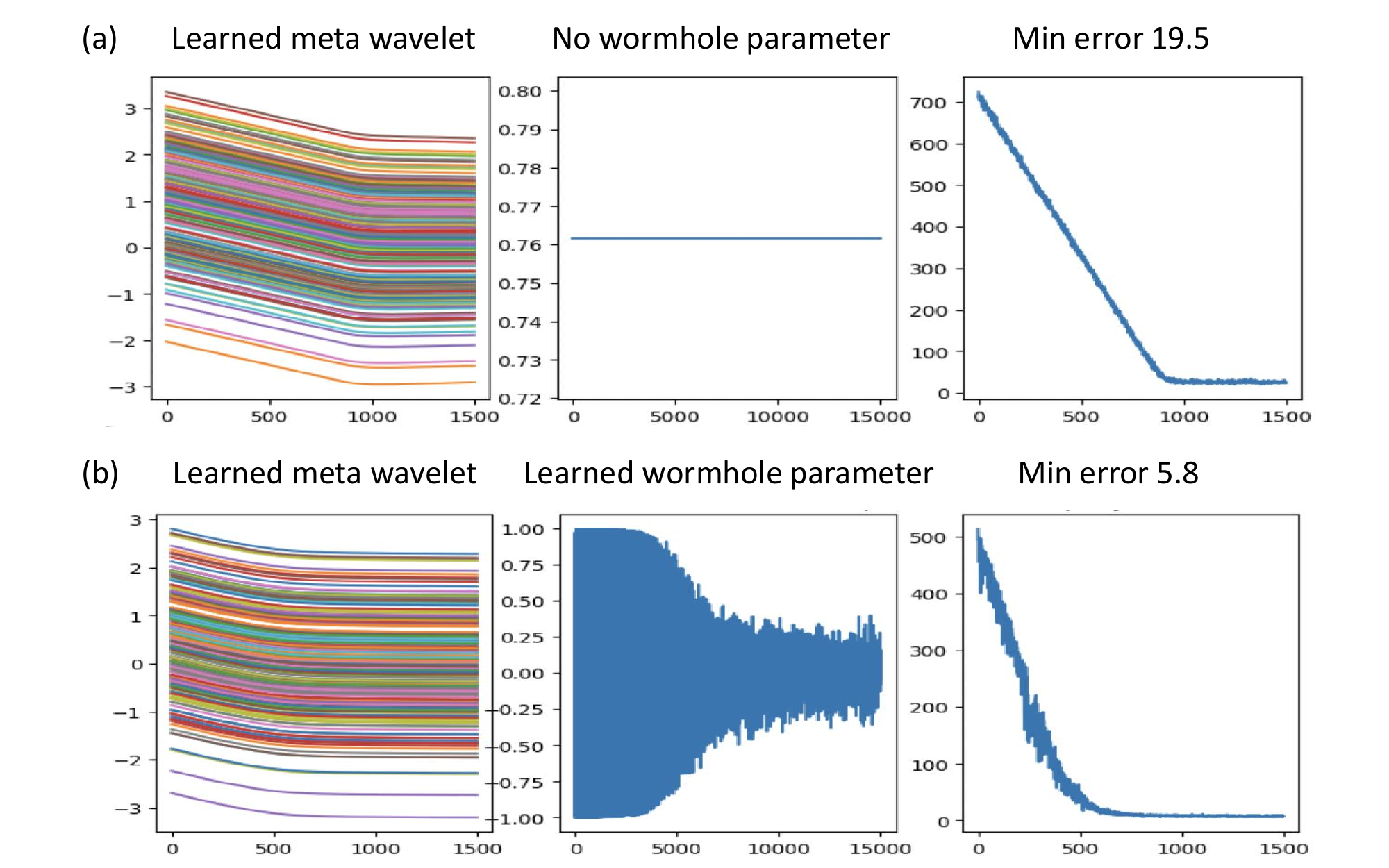}
\end{center}
\caption{\textbf{Parameter and loss curve analysis on the Wavelet Transform task.} (a) Without the adaptation of the wormhole parameter, the learned meta wavelet centers toward 0. (b) With the wormhole parameter learning sign and scale, the learned meta wavelet only needs to learn the shape.}
\label{fig:loss_wavelet}
\end{figure}
\paragraph{Results and Analysis}
Ideally, the meta $w$ would be in a bump-shape ranging in $[-0.4, 0.8]$, or $[-0.8, 0.4]$ which encompasses all wavelets under a single sign, since with $C_i$ as a sign changer $w$ only needs to learn the shape of the bump. However, without introducing $C_i$, the meta $w$ would simply center around 0 because the conflicting gradients of each task pull the values around zero. We further show that our results conform with the previous analysis in Fig.~\ref{fig:loss_wavelet}. Here for the sake of demonstration, we initialize $w$ randomly from $-2$ to $3$, and the inner loop adaptation for $C_i$ is 10 and for $w$ is $2$. As shown in the figure, without the adaptation of the wormhole parameter, the learned meta wavelet centers toward 0 at iteration 1500, while with the wormhole parameter learning sign and scale, the learned meta wavelet only needs to learn the shape, because at iteration 1500 the distribution of $w_i$ clearly shifts pass zero. This leads to a significant drop in the L1 error as well.

\subsection{MNIST Dataset}

\paragraph{Problem Formulation}
We consider a 2-way 1-shot classification on the MNIST dataset~\cite{lecun1998mnist}. Under the setting of 2-way classification, labels for the same digit class could be the complete opposite across different tasks, contributing to conflicting gradients during training.

\paragraph{Implementation Details}
For the MNIST dataset~\cite{lecun1998mnist}, we use a model that consists of two $5 \times 5$ convolutions followed by a ReLU nonlinearity and 2 × 2 max-pooling, and two linear layers with a ReLU nonlinearity. We train the model for 100 epochs with a task batch size of 16. The wormhole parameter $C_i = \tanh(t), \; t \in \mathbb{R}^1$ is added to the last linear layer of the model. An Adam optimizer with a learning rate of 0.005 is used in each experiment. We adopt the cross-entropy error between the predicted and ground-truth label as the loss function. We consider 1-, 2-, and 5-step inner loop adaptation in the experiments. The adaptation learning rate is set to 0.01.

\paragraph{Results and Analysis}
The meta-test classification error rate of each experiment is reported in Table~\ref{tab:evaluation}. While no advantage is observed due to training instability under the setting of 1-step adaptation, Wormhole MAML improves the vanilla MAML~\cite{finn2017model} by a large margin in both the 2- and 5-step settings, achieving the error rate of 0.141 and 0.009 respectively. 


\section{Limitation and Future Work}
As mentioned in the result analysis, the optimality of the multiplicative parameter $C_i$ found during inner-loop adaptation relies on relatively reasonable model parameters. While in many practical scenarios, a reasonable meta-parameter $\theta$ can lead to stable training dynamics for $C_i$, in some cases, one might need to optimize over $C_i$ and $\Delta_i$ at the same time. This potentially complicates the update rule, where some additional design is needed. For example, an annealing scheme may be worth investigating where $C_i$ converges faster than $\Delta_i$. In that case, the benefits of Wormhole MAML may slightly diminish due to the hassle of hyperparameter tuning.

It remains unclear how scalable Wormhole MAML is on more general problem sets. In more complicated problem settings with higher-dimensional models, the gradients are less likely to be completely contradictory, thus lessening the benefits of Wormhole MAML.

In addition, the choice of the form of $C_i$ is worth optimizing. A smaller dimension of $C_i$ ensures reliable and efficient optimization during each inner loop, but in some cases the expressivity may not be enough. A natural generalization of our approach would be to have $C_i$ in the same dimension as $\theta$, but constrained to be within a low-dimensional manifold, e.g. by a generative model. Taking one step further, one could constrain $\phi_i$ to a low-dimensional manifold. One potential problem with generalizing is the difficulty in propagating gradient through the inner-loop adaptation. More investigation is needed to offer insights on meta-learning in such glued parameter space.

\bibliography{iclr2022_conference}

\begin{thebibliography}{13}
\providecommand{\natexlab}[1]{#1}
\providecommand{\url}[1]{\texttt{#1}}
\expandafter\ifx\csname urlstyle\endcsname\relax
  \providecommand{\doi}[1]{doi: #1}\else
  \providecommand{\doi}{doi: \begingroup \urlstyle{rm}\Url}\fi

\bibitem[Duan et~al.(2017)Duan, Schulman, Chen, Bartlett, Sutskever, and
  Abbeel]{duan2016rl}
Yan Duan, John Schulman, Xi~Chen, Peter~L. Bartlett, Ilya Sutskever, and Pieter
  Abbeel.
\newblock {RL}{\textasciicircum}2: Fast reinforcement learning via slow
  reinforcement learning, 2017.
\newblock URL \url{https://openreview.net/forum?id=HkLXCE9lx}.

\bibitem[Finn et~al.(2017)Finn, Abbeel, and Levine]{finn2017model}
Chelsea Finn, Pieter Abbeel, and Sergey Levine.
\newblock Model-agnostic meta-learning for fast adaptation of deep networks.
\newblock In \emph{International conference on machine learning}, pp.\
  1126--1135. PMLR, 2017.

\bibitem[Flennerhag et~al.(2019)Flennerhag, Rusu, Pascanu, Visin, Yin, and
  Hadsell]{flennerhag2019meta}
Sebastian Flennerhag, Andrei~A Rusu, Razvan Pascanu, Francesco Visin, Hujun
  Yin, and Raia Hadsell.
\newblock Meta-learning with warped gradient descent.
\newblock \emph{arXiv preprint arXiv:1909.00025}, 2019.

\bibitem[LeCun(1998)]{lecun1998mnist}
Yann LeCun.
\newblock The mnist database of handwritten digits.
\newblock \emph{http://yann. lecun. com/exdb/mnist/}, 1998.

\bibitem[Lee \& Choi(2018)Lee and Choi]{mt-net}
Yoonho Lee and Seungjin Choi.
\newblock Gradient-based meta-learning with learned layerwise metric and
  subspace.
\newblock In Jennifer Dy and Andreas Krause (eds.), \emph{Proceedings of the
  35th International Conference on Machine Learning}, volume~80 of
  \emph{Proceedings of Machine Learning Research}, pp.\  2927--2936. PMLR,
  10--15 Jul 2018.

\bibitem[Li et~al.(2017)Li, Zhou, Chen, and Li]{li2017meta}
Zhenguo Li, Fengwei Zhou, Fei Chen, and Hang Li.
\newblock Meta-sgd: Learning to learn quickly for few-shot learning.
\newblock \emph{arXiv preprint arXiv:1707.09835}, 2017.

\bibitem[Lou et~al.(2021)Lou, Zhao, and Wang]{lou2021}
Beicheng Lou, Nathan Zhao, and Jiahui Wang.
\newblock Meta-learning from sparse recovery.
\newblock In \emph{Fifth Workshop on Meta-Learning at the Conference on Neural
  Information Processing Systems}, 2021.

\bibitem[Rajendran et~al.(2020)Rajendran, Irpan, and Jang]{rajendran2020meta}
Janarthanan Rajendran, Alexander Irpan, and Eric Jang.
\newblock Meta-learning requires meta-augmentation.
\newblock \emph{Advances in Neural Information Processing Systems},
  33:\penalty0 5705--5715, 2020.

\bibitem[Snell et~al.(2017)Snell, Swersky, and Zemel]{snell2017prototypical}
Jake Snell, Kevin Swersky, and Richard Zemel.
\newblock Prototypical networks for few-shot learning.
\newblock \emph{Advances in neural information processing systems}, 30, 2017.

\bibitem[Sun \& Saenko(2016)Sun and Saenko]{sun2016deep}
Baochen Sun and Kate Saenko.
\newblock Deep coral: Correlation alignment for deep domain adaptation.
\newblock In \emph{European conference on computer vision}, pp.\  443--450.
  Springer, 2016.

\bibitem[Yu et~al.(2020)Yu, Kumar, Gupta, Levine, Hausman, and
  Finn]{yu2020gradient}
Tianhe Yu, Saurabh Kumar, Abhishek Gupta, Sergey Levine, Karol Hausman, and
  Chelsea Finn.
\newblock Gradient surgery for multi-task learning.
\newblock \emph{Advances in Neural Information Processing Systems},
  33:\penalty0 5824--5836, 2020.

\bibitem[Zhang et~al.(2020)Zhang, He, Sra, and Jadbabaie]{gradient-clipping}
Jingzhao Zhang, Tianxing He, Suvrit Sra, and Ali Jadbabaie.
\newblock Why gradient clipping accelerates training: A theoretical
  justification for adaptivity.
\newblock In \emph{International Conference on Learning Representations}, 2020.

\bibitem[Zintgraf et~al.(2019)Zintgraf, Shiarli, Kurin, Hofmann, and
  Whiteson]{zintgraf2019fast}
Luisa Zintgraf, Kyriacos Shiarli, Vitaly Kurin, Katja Hofmann, and Shimon
  Whiteson.
\newblock Fast context adaptation via meta-learning.
\newblock In \emph{International Conference on Machine Learning}, pp.\
  7693--7702. PMLR, 2019.

\end{thebibliography}
\bibliographystyle{iclr2022_conference}

\end{document}